\documentclass[review]{elsarticle}
\graphicspath{{figures/}}

{}
{}
{}

\usepackage{lineno,hyperref}
\usepackage{multirow}
\usepackage{color}
\usepackage{makecell}
\usepackage{booktabs}
\begin{document}

\begin{frontmatter}

\title{Multi-focus Noisy Image Fusion using Low-Rank Representation}

\author[mymainaddress]{Hui Li}


\author[mymainaddress]{Xiao\_jun Wu\corref{mycorrespondingauthor}}
\cortext[mycorrespondingauthor]{Corresponding author}
\ead{xiaojun\_wu\_jun@163.com}

\address[mymainaddress]{School of Internet of Things Engineering, Jiangnan University,Wuxi 214122, China.}

\begin{abstract}
In the process of image acquisition, the noise is inevitable for source image. The multi-focus noisy image fusion is a very challenging task. However, there is no truly adaptive noisy image fusion approaches at present. As we all know, Low-Rank representation(LRR) is robust to noise and outliers. In this paper, we propose a novel fusion method based on LRR for multi-focus noisy image fusion. In the discrete wavelet transform(DWT) framework, the low frequency coefficients are fused by spatial frequency, the high frequency coefficients are fused by LRR coefficients and choose-max strategy. Finally, the fused image is obtained by inverse DWT. Experimental results demonstrate that the proposed algorithm can obtain state-of-the-art performance when the source images contain noise. The Code of our fusion method is available at \url{https://github.com/hli1221/imagefusion_noisy_lrr}.
\end{abstract}

\begin{keyword}
\texttt{multi-focus image fusion}\sep {low-rank representation} \sep {noisy image fusion}
\end{keyword}

\end{frontmatter}

\section{Introduction}\label{sec1}

Multi-focus image fusion is an important technique in image processing field. The main purpose of image fusion is to obtain a single image by fusing complementary information from source images[1].Image fusion methods can be divided into two categories: non-representation learning-based methods and representation learning-based methods. In representation learning-based methods, the input images or features are mapped into another domain which makes the problem to be solved easily. And adaptive strategies are utilized to fuse salient features [19][20][21]. Then, fused images will be obtained by these features and input images. In contrast with representation learning-based methods, we called non-representation learning-based methods.

In non-representation learning-based fusion methods, multi-scale transforms are the most commonly fusion methods, such as discrete wavelet transform(DWT)[2], contourlet[3] and shearlet[4][18]. Due to the wavelet transform has not enough detail preservation ability, in reference[5], non-sampled contourlet transform(NSCT) was applied to image fusion.

In addition, the morphology which is also a non-representation learning technique was applied to image fusion. Zhang et al.[6] proposed a fusion method based on morphological gradient. The detail information (like texture and edge) is obtained by different morphological gradient operators. Then the boundary region of focus and defocus, focus region and defocus region are extracted by this information. Finally, the fused image can be obtained by an appropriate fusion strategy.

In representation learning-based fusion methods, the convolutional neural network(CNN)[22], low-rank representation(LRR), and sparse representation(SR) techniques have various applications in image processing.
 
Liu et al.[7] proposed the CNN-based image fusion methods. A decision map is obtained by the output of CNN which is trained by image patches and different blurred version. Finally, the fused image is obtained by the decision map. However, this method needs lot of images to train the network, even when the images are divided into patches.

For the first time, LRR is applied to image fusion tasks by Li et al.[11]. In their algorithm, K-singular value decomposition (K-SVD) is used to calculate a global dictionary which is utilized to obtained low-rank coefficients of source images. Then, $l_1$-norm and choose-max strategy are used to fuse these coefficients. Finally, fused image is reconstructed by global dictionary and fused low-rank coefficients. Due to the dictionary learning is applied to LRR, time efficiency of this methods is very low.

The sparse representation(SR) method[8], [9] is a classical technique in representation learning-based methods. SR-based image fusion has great performance in some image fusion tasks, but it still suffers from many drawbacks: 1) It is difficult to learn a good dictionary in offline manner; 2) The time efficiency of SR-based methods is very low, especially when using online manner to learn dictionary. Due to these drawbacks, Liu et al.[10] proposed a novel image fusion method based on convolutional sparse representation (ConvSR). The CSR-based fusion method obtained better fusion performance.

Besides the above drawbacks, the SR-based image fusion method cannot capture the global structure of image. Furthermore, when the source images contain noise, the image fusion performance obtained by above fusion methods will become worse.

In order to address these problems, we apply LRR to multi-focus image fusion task. As we all know, the LRR is robust to noise and outliers[12]. So the LRR technique is a prefect tool for multi-focus noisy image fusion. In this study, we propose a novel multi-focus image fusion method based on LRR in noisy image fusion task and this method will be introduced in the next section.
 
The rest of this paper is organized as follows. In Section 2, we introduce the LRR theory briefly. In Section 3, the proposed LRR-based image fusion method will be introduced in detail. The experimental setting and fusion results are provided in Section 4. Section 5 draws the conclusions.

\begin{figure}[ht]
\centering
\includegraphics[width=\linewidth]{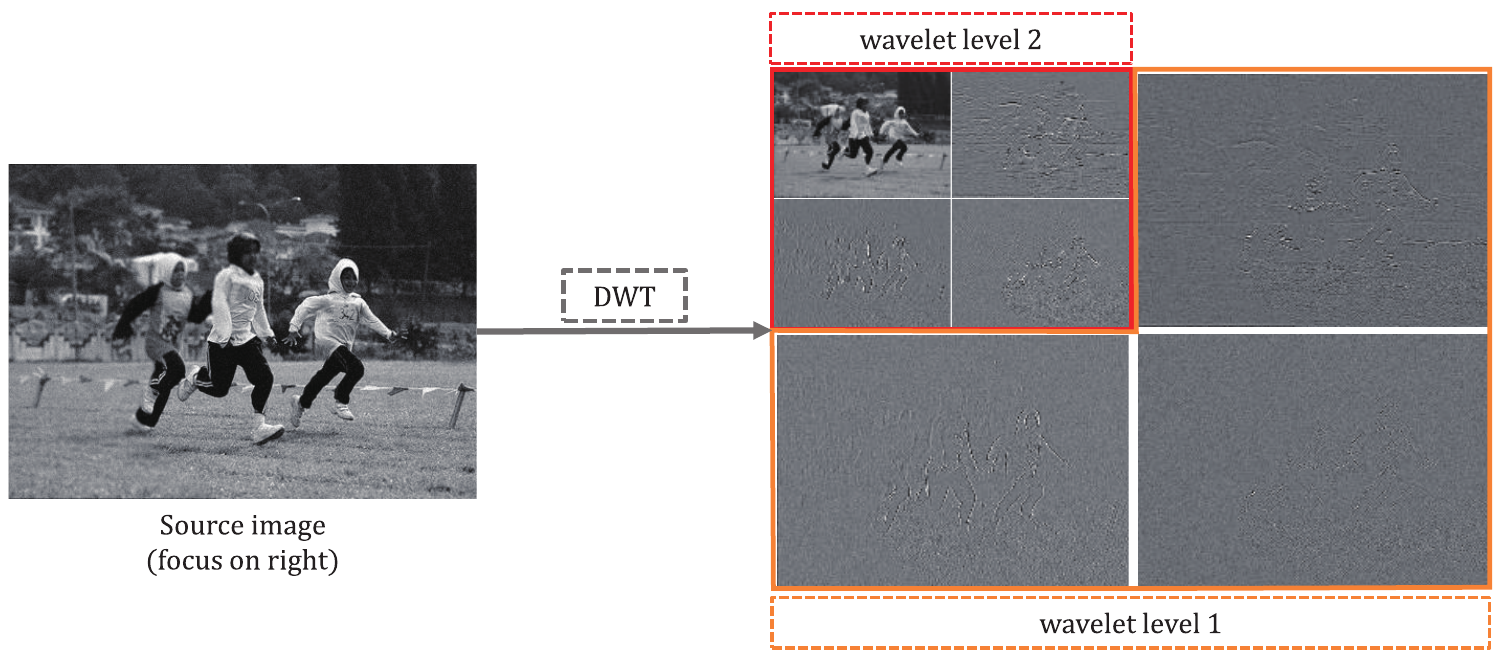}
\caption{The procedure of DWT operation.}
\label{fig:fig1}
\end{figure}

\section{Related work}\label{sec2}

\textbf{Discrete Wavelet Transform(DWT).} In image fusion tasks, DWT[2] is a classical and useful technique for image processing. With DWT operation, input images are decomposed into several coefficient matrices which are low frequency band and high frequency band. The size of coefficient matrices will reduce with increase of the level of wavelet decomposition.

In our paper, the procedure of DWT operation is shown in Fig.\ref{fig:fig1}. 

In Fig.\ref{fig:fig1}, the decomposition level of DWT is 2. After DWT operation, seven matrices are obtained. `wavelet level 1' contains three high frequency matrices(upper right: feature of horizontal; bottom left: feature of vertical; bottom right: feature of diagonal). `wavelet level 2' contains one low frequency matrix(upper left) and three high frequency matrices, in which the feature orientations are the same as`wavelet level 1'.

\textbf{Low-rank Representation(LRR).} In order to capture the global structure of data, Liu et al.[12] proposed a novel representation method, namely, low-rank representation(LRR).

In reference [12], authors apply self-expression model to avoid training a dictionary and the LRR problem is solved by the following optimization problem,
\begin{eqnarray}\label{equ:1}
  	\min_{Z,E}||Z||_*+\lambda||E||_{2,1} \\
    s.t.,X=XZ+E \nonumber
\end{eqnarray}
where $X$ denotes the observed data matrix, $E$ indicates the noise matrix, $||\cdot||_*$ denotes the nuclear norm which is the sum of the singular values of matrix. $||E||_{2,1}=\sum_{j=1}^{n}\sqrt{\sum_{i=1}^{n}[E]_{ij}^2}$ is called as $l_{2,1}$-norm, $\lambda>0$ is the balance coefficient. Eq.\ref{equ:1} is solved by the inexact Augmented Lagrange Multiplier (ALM). Finally, the LRR coefficients matrix $Z$ for $X$ is obtained by Eq.\ref{equ:1}.

\begin{figure*}[!ht]
\centering
\includegraphics[width=\linewidth]{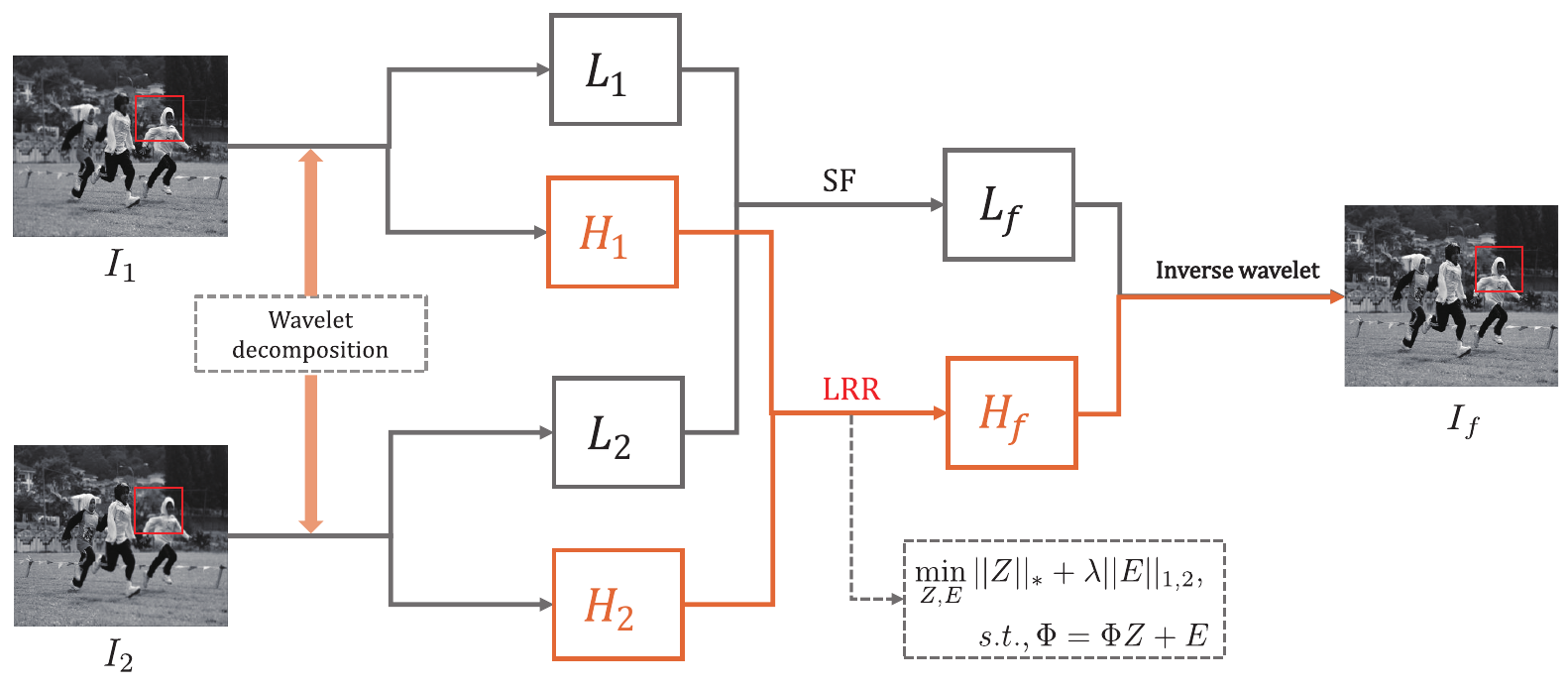}
\caption{The framework of the proposed method.}
\label{fig:fig2}
\end{figure*}

\section{The Proposed Image Fusion method}\label{sec3}

In this section, we intend to propose a novel method based on LRR theory in DWT domain. In this paper, the source images(focus on right and left) are denoted as $I_1$ and $I_2$. Note that the fusion strategy is the same when the input images more than 2. And the indices(1, 2) are irrelevant with the focus type. The system diagram of our proposed method is shown in Fig.\ref{fig:fig2}.

Firstly, the source images $I_1$ and $I_2$ are decomposed by DWT operation. We choose the decomposition level of DWT as 2. 

Then, the spatial frequency (SF) and choose-max strategy are used to fuse the low frequency coefficients since the low frequency coefficients reflect the non-detail information of source images. The high frequency coefficients include more detail information of source image, so we choose the LRR to get a low rank matrix and use the nuclear norm and choose-max scheme to fuse the high frequency coefficients. In Fig.\ref{fig:fig2} (LRR), $\Phi$ indicates the input high frequency coefficients.

Finally, the fused image is obtained by inverse DWT operation.

\subsection{Fusion of low frequency coefficients}

The low frequency coefficients contain more contour information and less detail texture information. Thus, the spatial frequency(SF)[7] is used to fuse low frequency coefficients. The SF is calculated by Eq.\ref{equ:2} - \ref{equ:4},
\begin{eqnarray}\label{equ:2}
	SF =& \sqrt{f_{x}^{2}+f_{y}^{2}} \\
    \label{equ:3}
    f_{x} =& \sqrt{\frac{1}{MN}\sum_{i=0}^{M-1}\sum_{j=1}^{N-1}[f(i,j)-f(i,j-1)]^{2}} \\
    \label{equ:4}
    f_{y} =& \sqrt{\frac{1}{MN}\sum_{i=1}^{M-1}\sum_{j=0}^{N-1}[f(i,j)-f(i-1,j)]^{2}}
\end{eqnarray}
\noindent where $f_x$ and $f_y$ are spatial frequency of $x$ and $y$ directions, $M$ and $N$ are the row and column numbers of the image.

By sliding window technique, the coefficient matrices are divided into $M_l$ patches. Then the SF value of adjacent coefficient patches are obtained by Eq.\ref{equ:2}-\ref{equ:4}. Finally, we use the choose-max scheme to get the fused low frequency coefficients.

Let $SF_K^r$ denote the SF value of each patch, where $K\in{\{1,2\}}$ denotes the SF value from which source images, and $r\in{\{1,\cdots,M_l\}}$ denotes $r$-th patch in source image. Thus, the fused low frequency coefficients $L_f$ is obtained by Eq.\ref{equ:5}.

\begin{equation}\label{equ:5}
	L_f^r=\left\{\begin{array}{ll}
		L_{1}^{r} & \textrm{ $SF_{1}^{r}>SF_{1}^{r}$ } \\
        L_{2}^{r} & \textrm{ otherwise }
	\end{array}\right.
\end{equation}

\subsection{Fusion of high frequency coefficients}

\begin{figure*}[ht]
\centering
\includegraphics[width=0.8\linewidth]{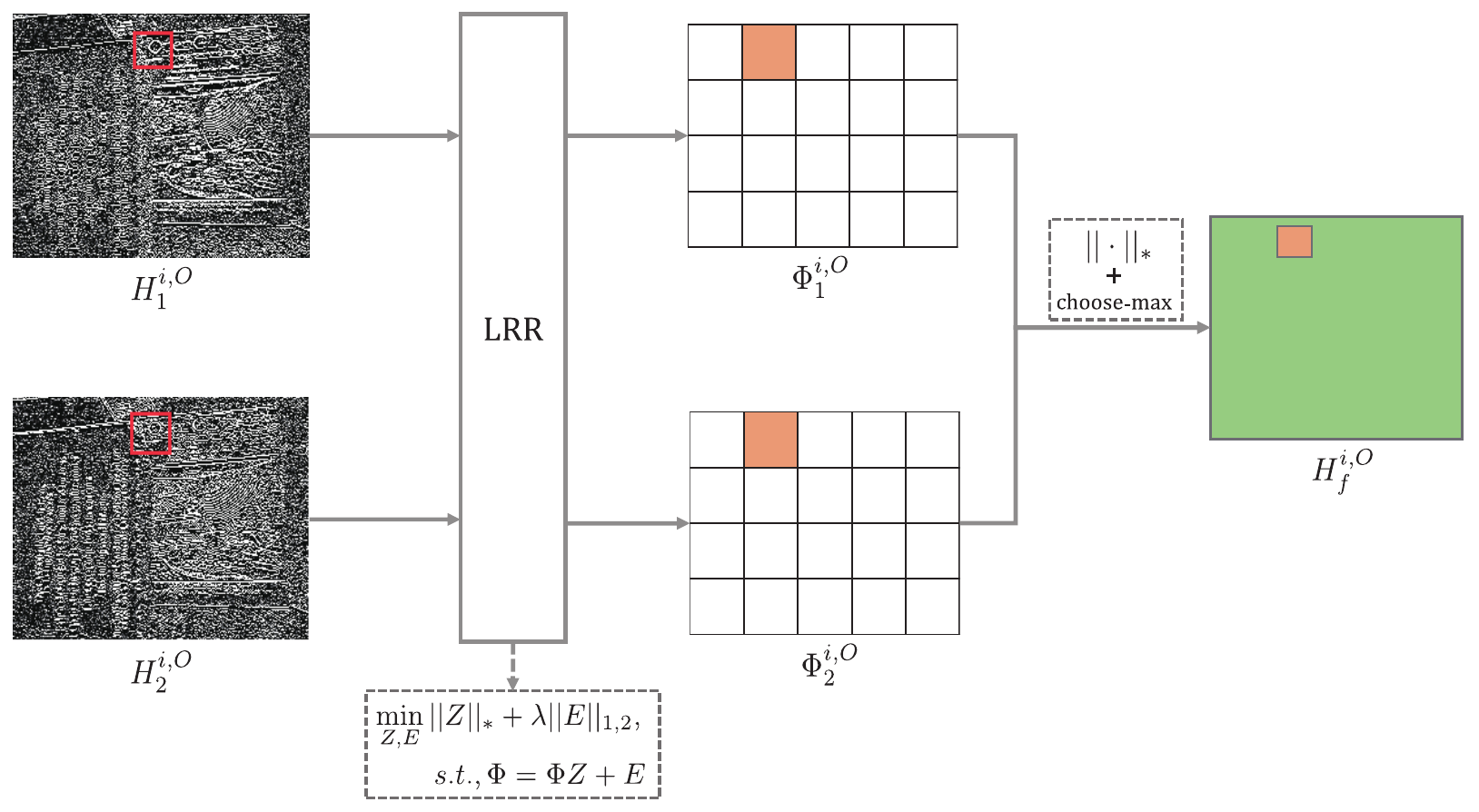}
\caption{The framework of the fusion of high frequency coefficient.}
\label{fig:fig3}
\end{figure*}

Let $H_K^{i,O}$ denote the high frequency coefficients which are obtained by the 2-level DWT, where $O\in{\{H,D,V\}}$ represents the direction of decomposition and $H$ stand for horizontal, $D$ for diagonal, $V$ for vertical, and $K\in{\{1,2\}}$ denotes the index of source images, $i\in{\{1,2\}}$ represents the level of DWT operation, each level has 3 high frequency coefficients matrices. Thus, we get 6 high frequency coefficients matrices and one low frequency matrix. The fusion strategy of high frequency coefficients are shown in Fig.\ref{fig:fig3}.

By sliding window technique, each high frequency coefficients matrix is divided into $M_h$ patches. The size of window is $n\times n$, $H_{K,j}^{i,O}$ denotes one image patch, $j\in{\{1,\cdots,M_h\}}$ denotes $j$-th patch, $i\in{\{1,2\}}$ represents the level of DWT, $O\in{\{H,D,V\}}$ indicates the direction of decomposition. Then two low-rank matrices $\Phi_{1,j}^{i,O}$ and $\Phi_{2,j}^{i,O}$ are obtained by the LRR theory. The local fused high frequency coefficients matrix $H_{f,j}^{i,O}$ is obtained by comparing the nuclear norm of corresponding low-rank coefficients $Z$.

Suppose the local high frequency coefficients matrix is denoted by $\Phi_{K,j}^{i,O}$. Applying the self-expression model, and we use $\Phi_{K,j}^{i,O}$ itself as the dictionary. The Eq.\ref{equ:6} is used to obtain the low-rank matrix.
\begin{eqnarray}\label{equ:6}
  	\min_{Z_{K,j}^{i,O},E}||Z_{K,j}^{i,O}||_*+\lambda||E||_{2,1} \\
    s.t. \Phi_{K,j}^{i,O}=\Phi_{K,j}^{i,O} Z_{K,j}^{i,O}+E \nonumber
\end{eqnarray}

We choose the inexact ALM to solve the problem \ref{equ:6}. Then the low-rank coefficients matrix $Z_{K,j}^{i,O}$ and the noise matrix $E$ are obtained. Note that $E$ is noise matrix, so $E$ is ignored in this step. The nuclear norm $||Z_{K,j}^{i,O}||_*$ is obtained by computing the sum of the singular values of the matrix $Z_{K,j}^{i,O}$. Finally, the fused high frequency coefficients matrices are calculated by Eq.\ref{equ:7},
\begin{equation}\label{equ:7}
	H_{f,j}^{i,O}=\left\{\begin{array}{ll}
		\Phi_{1,j}^{i,O} & \textrm{ $||Z_{1,j}^{i,O}||_*>||Z_{2,j}^{i,O}||_*$ } \\
        \Phi_{2,j}^{i,O} & \textrm{ otherwise }
	\end{array}\right.
\end{equation}

\noindent where $\Phi_{K,j}^{i,O}=\Phi_{K,j}^{i,O} Z_{K,j}^{i,O}$ denotes the low-rank representation coefficients of $H_{K,j}^{i,O}$. 

\subsection{Reconstruction of fused image}
Having the fused coefficients $L_f$ and $H_f$, the fused image $I_f$ is reconstructed by inverse DWT.
\begin{eqnarray}\label{equ:8}
  	I_f = IDWT(L_f,H_f)
\end{eqnarray}

\noindent In Eq.\ref{equ:8}, $IDWT(\cdot)$ indicates the inverse DWT operation.

The procedure of our method is described as follows.

	1) The source images are decomposed by 2-level DWT. Then the low frequency coefficients $L_K$ and the high frequency coefficients $H_K^{i,O}$ are obtained, where $K\in{\{1,2\}}$(source images), $O\in{\{H,D,V\}}$(direction of decomposition), $i\in{\{1,2\}}$(level of DWT).

	2) By sliding window technique, the low frequency coefficients are divided into $M_l$ patches and the high frequency coefficients are divided into $M_h$ patches.

	3) For low frequency coefficients, we use SF and choose-max scheme to fuse these coefficients.

	4) For high frequency coefficients, LRR is used to compute the low-rank matrix $Z$. Then, $||Z||_*$ and choose-max scheme are utilized to get fused high frequency coefficients.

	5) Finally, with the fused low frequency coefficients and high frequency coefficients, the fused image is obtained by inverse DWT

\section{Experiment}\label{sec4}

In this section, firstly, we introduce our experimental database. And how to choose the LRR parameter($\lambda$), image patch size and wavelet level in different situations(different noises) are shown in Section \ref{para} and \ref{size_level}. 

Then, we introduce the detail experimental settings and analyze the fused results. Finally, the fusion results are shown in the last section. 

The experiments are implemented in MTALAB R2016a on 3.2 GHz Intel(R) Core(TM) CPU with 12 GB RAM.

\subsection{Experimental data}
\begin{figure*}[ht]
\centering
\includegraphics[width=0.9\linewidth]{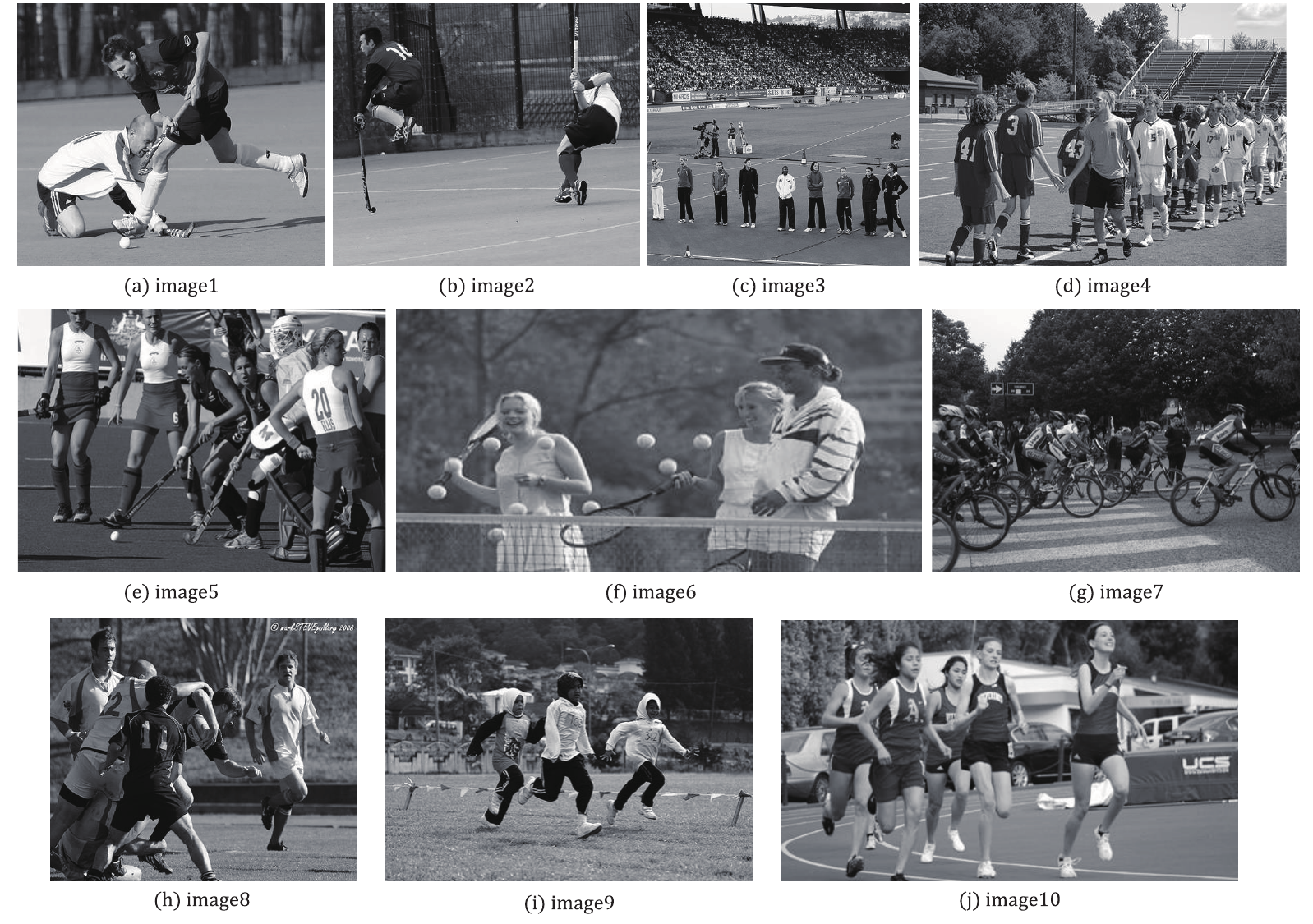}
\caption{Ten images form ImageNet. In next section, we use image1-10 represent these images.}
\label{fig:fig4}
\end{figure*}

In our experiment, we choose ten images from ImageNet in sport(\url{http://www.image-net.org/index}), as shown in Fig.\ref{fig:fig4}. We blur these images to get source images. Gaussian smoothing filter with size $3\times 3$ and $\sigma=7$ is used to blur these images. 

Furthermore, the noise of Gaussian ($\mu=0,\sigma=0.0005, 0.001, \\0.005, 0.01$), salt \& pepper (the noise density is 0.01 and 0.02) and the noise of Poisson are utilized to process these blurred version images. Then we use these images to determine the parameter($\lambda$) in different situations and compare the results of fusion methods.

\begin{figure*}[ht]
\centering
\includegraphics[width=0.9\linewidth]{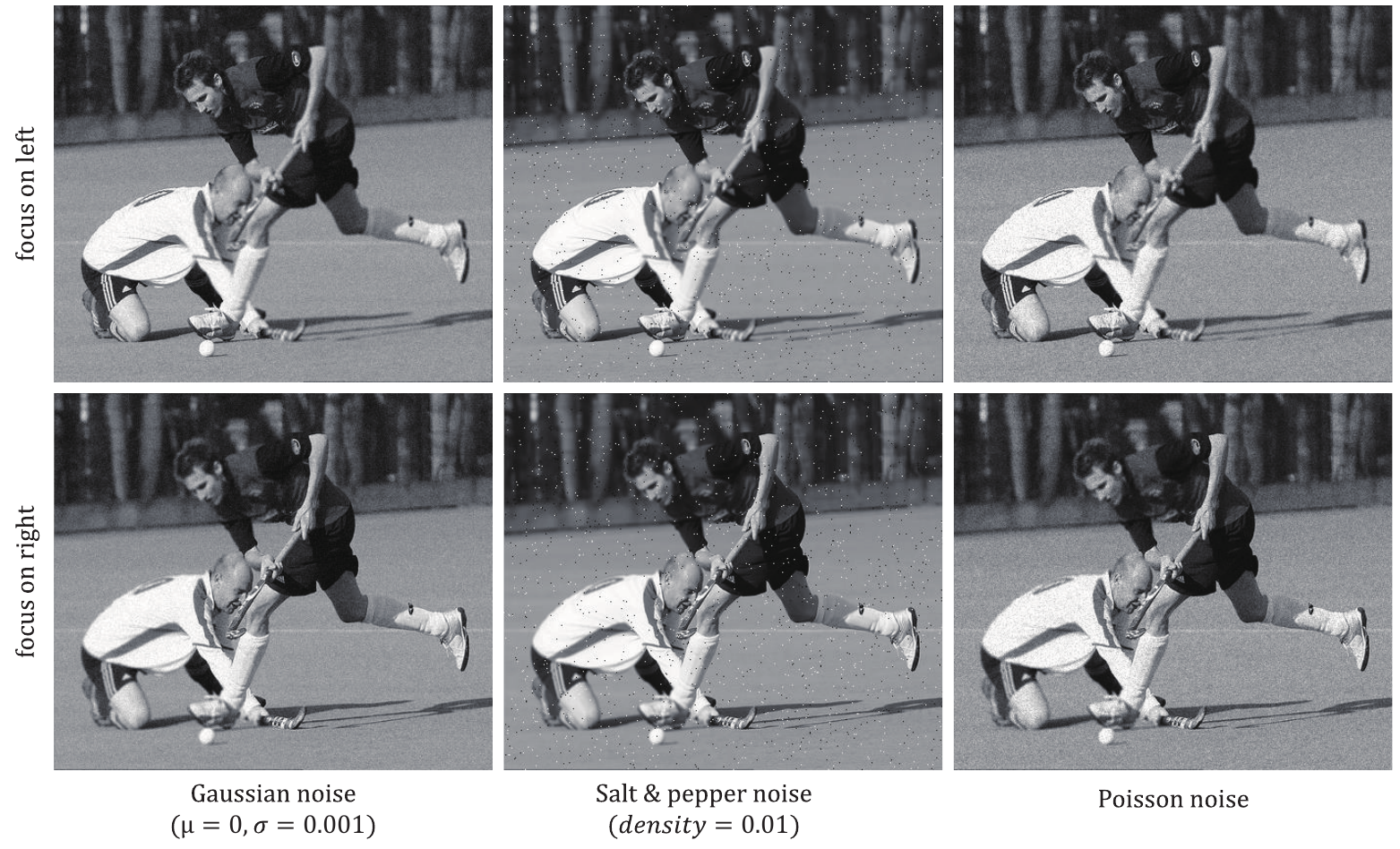}
\caption{Example for different noise (image1).}
\label{fig:fig5}
\end{figure*}

The ``image1'' contains three different noise and focus on different regions are shown in Fig.\ref{fig:fig5}.

In order to evaluate the proposed method and other fusion methods, three quality metrics are utilized. These are: Root Mean Square Error(RMSE), Peak Signal to Noise Ratio (PSNR) and Structural Similarity (SSIM). In particular, RMSE, PSNR and SSIM are based on reference image. The fused image is better when the values of PSNR and SSIM are larger. On the contrary, the fused image is more similar to reference image when the values of RMSE is smaller.

\subsection{Effects of parameter}\label{para}

In Eq.\ref{equ:6}, the parameter $\lambda>0$ is used to balance the effects of the low rank part($Z$) and noise part($E$). In this section, we choose image1-5(Fig.\ref{fig:fig4} a-e) and their blurry versions which contain three different types of noise as the source images to determine the parameter $\lambda$ in our fusion framework.

And in this section, the image patch size and wavelet level are set as $16\times 16$ and 2, respectively. And the settings for patch size and wavelet level are only temporary, we will discuss how to choose the size of image patch and the level of wavelet in next section.

The range of $\lambda$ is set as [1,50] for Gaussian noise($\mu=0,\sigma=0.0005, 0.001, 0.005, 0.01$) and salt \& pepper noise(noise density is 0.01 and 0.02), and the range of $\lambda$ is set as [1,20] for Poisson noise. The different situations include Gaussian noise, Salt \& pepper noise and Poisson noise. We choose the average SSIM value to determine the parameter $\lambda$ which will be used in fusion phase.
\begin{figure*}[ht]
\centering
\includegraphics[width=0.9\linewidth]{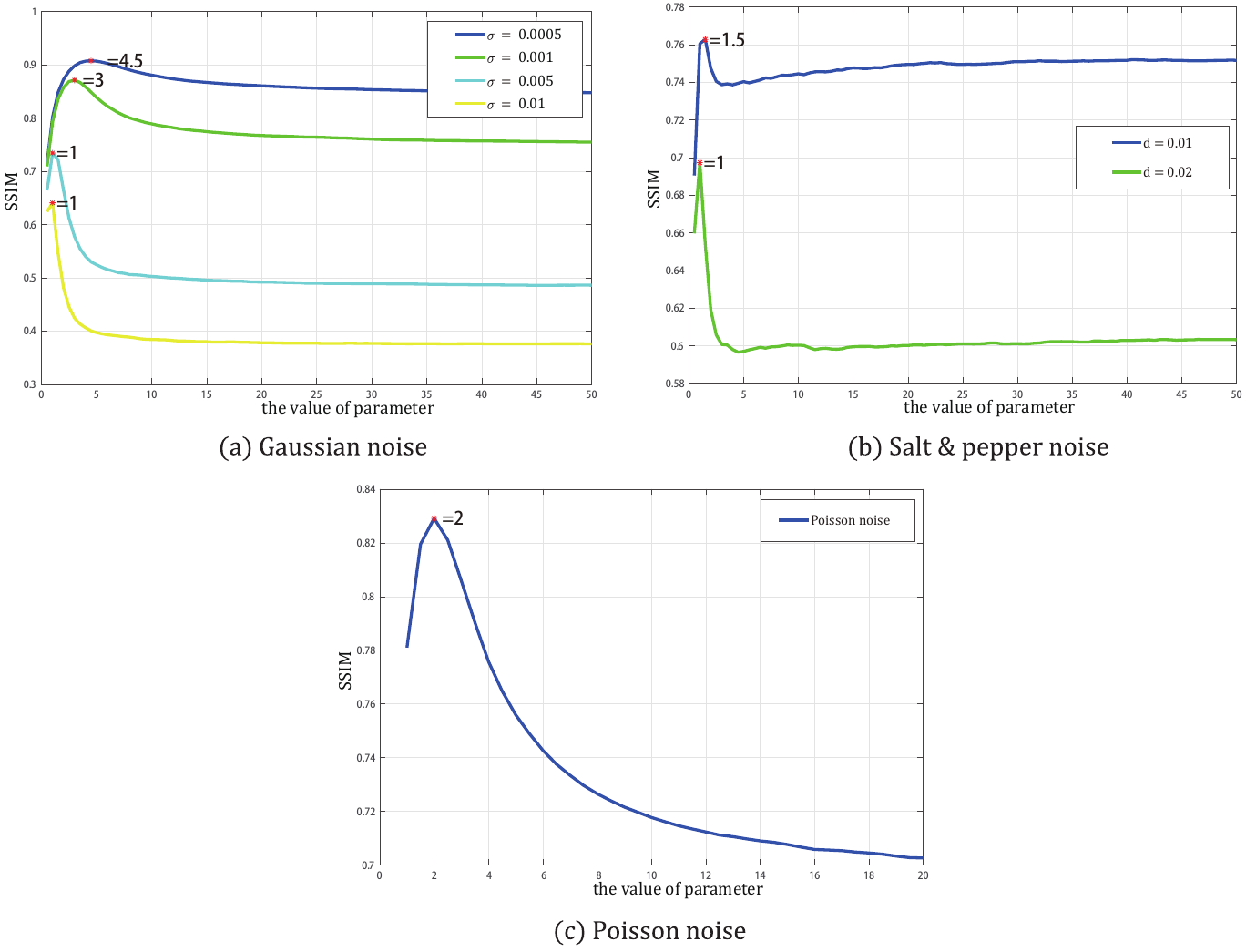}
\caption{Average SSIM values with different noise. (a) Gaussian noise with different $\sigma$; (b) Salt \$ pepper noise with different density; (c) Poisson noise.}
\label{fig:fig6}
\end{figure*}

As shown in Fig.\ref{fig:fig6}, when the source images contain Gaussian noise(a), with the increase of $sigma$, the parameter $\lambda$ will reduce to separate more noise part from input matrix. And SSIM will get maximum value at $\lambda=4.5,3,1,1$ when $sigma=0.0005,0.001,0.005,0.01$, respectively.

And the trend is the same for Salt \& pepper noise in which the noise density is set as 0.01 and 0.02. SSIM will get maximum value at $\lambda=1.5,1$ when $d=0.01,0.02$, respectively.
When source images contain Poisson noise, $\lambda$ is set as 2 to get the maximum SSIM value.

Therefore, we choose $\lambda=4.5,3,1,1$ for containing Gaussian noise in which $\sigma=0.0005,0.001,0.005,0.01$, respectively. $\lambda=1.5,1$ when the source images contain salt \& pepper noise($d=0.01,0.02$), and $\lambda=2$ for Poisson noise.

\subsection{Patch size and wavelet level}\label{size_level}

Once the parameter $\lambda$ is fixed for different noises, we change the image patch size and wavelet level to choose the best patch size and level in our fusion framework. The LRR parameter $\lambda$ is set as we discussed in Section ref{para}.

In this section, we calculate the average values of RMSE, PSNR and SSIM for ten pairs of source images. We choose three levels of wavelet, named level 1, level 2, level 3. And four types of images patch size($4\times 4, 8\times 8, 16\times 16, 32\times 32$). The average values of RMSE, PSNR and SSIM are shown in Table \ref{table:1}. The best values are denoted in bold and the second-best values are indicated in red.

\begin{table*}[ht]
\centering
\caption{\label{table:1}The average values of RMSE, PSNR and SSIM with different patch size and wavelet levels.}
\resizebox{\textwidth}{!}{
\begin{tabular}{*{15}{c}}
\midrule
\multirow{2}*{Noise} & \multirow{2}*{-} &\multirow{2}*{Metrics} &\multicolumn{4}{c}{level 1} &\multicolumn{4}{c}{level 2} &\multicolumn{4}{c}{level 3}\\
\cmidrule(lr){4-7} \cmidrule(lr){8-11} \cmidrule(lr){12-15}
 & & &$4\times 4$ &$8\times 8$ &$16\times 16$ &$32\times 32$ &$4\times 4$ &$8\times 8$ &$16\times 16$ &$32\times 32$ &$4\times 4$ &$8\times 8$ &$16\times 16$ &$32\times 32$\\
\midrule
\multirow{12}*{{\shortstack{Gaussian\\ noise}}} &
\multirow{3}*{$\sigma=0.0005$} &
  RMSE	&0.04913 	&0.04913 	&0.04939 	&0.04979 	&0.02983 	&\color{red}{0.02534} 	&\textbf{0.02327} 	&0.03377 	&0.12103 	&0.11986 	&0.11951 	&0.12260\\ 
&&PSNR	&26.48348   &26.48923   &26.44441 	&26.37138 	&30.59645 	&\color{red}{31.98071} 	&\textbf{32.72188} 	&29.64172 	&18.47811 	&18.57138 	&18.60120 	&18.35893\\ 
&&SSIM	&0.81914 	&0.82137 	&0.81838 	&0.81049 	&0.87445 	&\color{red}{0.90230} 	&\textbf{0.91120} 	&0.86410 	&0.81078 	&0.84021 	&0.84660 	&0.80616\\ 
\cmidrule{2-15}
&\multirow{3}*{$\sigma=0.001$} &
  RMSE	&0.05033 	&0.05024 	&0.05053 	&0.05111 	&0.03639 	&\color{red}{0.03113} 	&\textbf{0.02839} 	&0.03733 	&0.12348 	&0.12149 	&0.12078 	&0.12367\\ 
&&PSNR	&26.24008   &26.26435   &26.21773 	&26.11500 	&28.89271 	&\color{red}{30.21484} 	&\textbf{30.99987} 	&28.72740 	&18.28901 	&18.44223 	&18.49942 	&18.27654\\ 
&&SSIM	&0.78905 	&0.79368 	&0.79110 	&0.78133 	&0.83254 	&\color{red}{0.86378} 	&\textbf{0.87428} 	&0.82320 	&0.76709 	&0.80377 	&0.81196 	&0.76728\\ 
\cmidrule{2-15}
&\multirow{3}*{$\sigma=0.005$} &
  RMSE	&0.06024 	&0.05942 	&0.05920 	&0.05959 	&0.06267 	&0.05625 	&\textbf{0.05059} 	&\color{red}{0.05209} 	&0.13737 	&0.13283 	&0.25661 	&0.13058\\ 
&&PSNR	&24.53899   &24.66591   &24.70275 	&24.64929 	&24.22779 	&25.13552 	&\textbf{26.02700} 	&\color{red}{25.78012} 	&17.30404 	&17.60930 	&11.92927 	&17.76885\\ 
&&SSIM	&0.63780 	&0.64995 	&0.65484 	&0.65078 	&0.66772 	&\color{red}{0.70601} 	&\textbf{0.73387} 	&0.70366 	&0.58684 	&0.64162 	&0.52215 	&0.64186\\ 
\cmidrule{2-15}
&\multirow{3}*{$\sigma=0.01$} &
  RMSE	&0.07059 	&0.06930 	&0.06938 	&0.07161 	&0.06537 	&\color{red}{0.05947} 	&\textbf{0.05714} 	&0.06597 	&0.13778 	&0.13354 	&0.25543 	&0.13507\\ 
&&PSNR	&23.10186   &23.26629   &23.25991 	&22.98052 	&23.83543 	&\color{red}{24.62688} 	&\textbf{24.92995} 	&23.66065 	&17.27658 	&17.55954 	&11.96730 	&17.46202\\ 
&&SSIM	&0.54179 	&0.55403 	&0.55536 	&0.53552 	&0.63020 	&\textbf{0.66433} 	&0.64582 	&0.55248 	&0.58010 	&0.62457 	&0.46971 	&0.51068\\ 
\midrule

\multirow{6}*{{\shortstack{Salt \& pepper\\ noise}}} &
\multirow{3}*{$d=0.01$} &
  RMSE	&0.06027 	&0.05873 	&0.05794 	&0.05799 	&0.05652 	&\color{red}{0.05195} 	&\textbf{0.04882} 	&0.05275 	&0.13255 	&0.12944 	&0.12790 	&0.12970\\ 
&&PSNR	&24.54166   &24.78140   &24.91218 	&24.91224 	&25.07708 	&\color{red}{25.77278} 	&\textbf{26.28470} 	&25.62336 	&17.62857 	&17.84646 	&17.95766 	&17.82983\\ 
&&SSIM	&0.71254 	&0.73231 	&0.74354 	&0.74463 	&0.70838 	&0.73976 	&\textbf{0.76127} 	&\color{red}{0.74540} 	&0.63481 	&0.67707 	&0.69568 	&0.68313\\ 
\cmidrule{2-15}
&\multirow{3}*{$d=0.02$} &
  RMSE	&0.06814 	&0.06546 	&0.06427 	&0.06465 	&0.06465 	&\color{red}{0.05943} 	&\textbf{0.05606} 	&0.05990 	&0.13792 	&0.13386 	&0.13203 	&0.13361\\ 
&&PSNR	&23.42172   &23.78491   &23.95458 	&23.90851 	&23.94129 	&\color{red}{24.63398} 	&\textbf{25.10640} 	&24.52367 	&17.26762 	&17.53855 	&17.66385 	&17.55841\\ 
&&SSIM	&0.63032 	&0.66192 	&0.67632 	&0.67273 	&0.65153 	&\color{red}{0.68369} 	&\textbf{0.69728} 	&0.66568 	&0.57734 	&0.62144 	&0.63058 	&0.60254\\ 
\midrule

\multirow{3}*{Poisson noise} &
\multirow{3}*{--} &
  RMSE	&0.05186 	&0.05161 	&0.05181 	&0.05241 	&0.04490 	&\color{red}{0.03874} 	&\textbf{0.03459} 	&0.04105 	&0.12726 	&0.12428 	&0.12282 	&0.12501\\ 
&&PSNR	&25.94134   &25.99243   &25.96176 	&25.86175 	&27.09106 	&\color{red}{28.33552} 	&\textbf{29.28811} 	&27.86859 	&18.00680 	&18.22849 	&18.34065 	&18.17565\\ 
&&SSIM	&0.76342 	&0.76972 	&0.76891 	&0.76051 	&0.78243 	&\color{red}{0.81914} 	&\textbf{0.83724} 	&0.79547 	&0.71025 	&0.75717 	&0.77362 	&0.73820\\ 
\hline
\end{tabular}}
\end{table*}

From Table \ref{table:1}, when the patch size and wavelet level are set as $16\times 16$ and 2, we will get almost the best and the second-best values for different noises. 

So in the next experiment, in our method, the wavelet level and image patch size are set as 2 and $16\times 16$, respectively. And the LRR parameter $\lambda$ is set as 4.5, 3, 1 and 1 for Gaussian noise($\mu=0,\sigma=0.0005, 0.001, 0.005, 0.01$), 1.5 and 1 for salt \& pepper noise($d=0.01,0.02$). $\lambda=2$ for Poisson noise.

\subsection{Fusion results}\label{results}

In this section, ten pairs of images (contain different noise for source images) are used to assess the performance of these methods numerically. 

The fusion performance of the proposed method is evaluated against other base line methods, we choose nine state-of-the-art existing methods, including: discrete wavelet transform (DWT)[2], cross bilateral filter fusion method (CBF)[13], discrete cosine harmonic wavelet transform fusion method(DCHWT)[14], multi-scale weighted gradient-based fusion method[15], weighted least square optimization(WLS)[16], convolutional sparse representation(ConvSR)[10], a deep convolutional neural network based fusion method(CNN)[7], multi-layers fusion method(MLVGG)[17] and LRR with dictionary learning based fusion method(DLLRR)[11].

\begin{figure*}[ht]
\centering
\includegraphics[width=\linewidth]{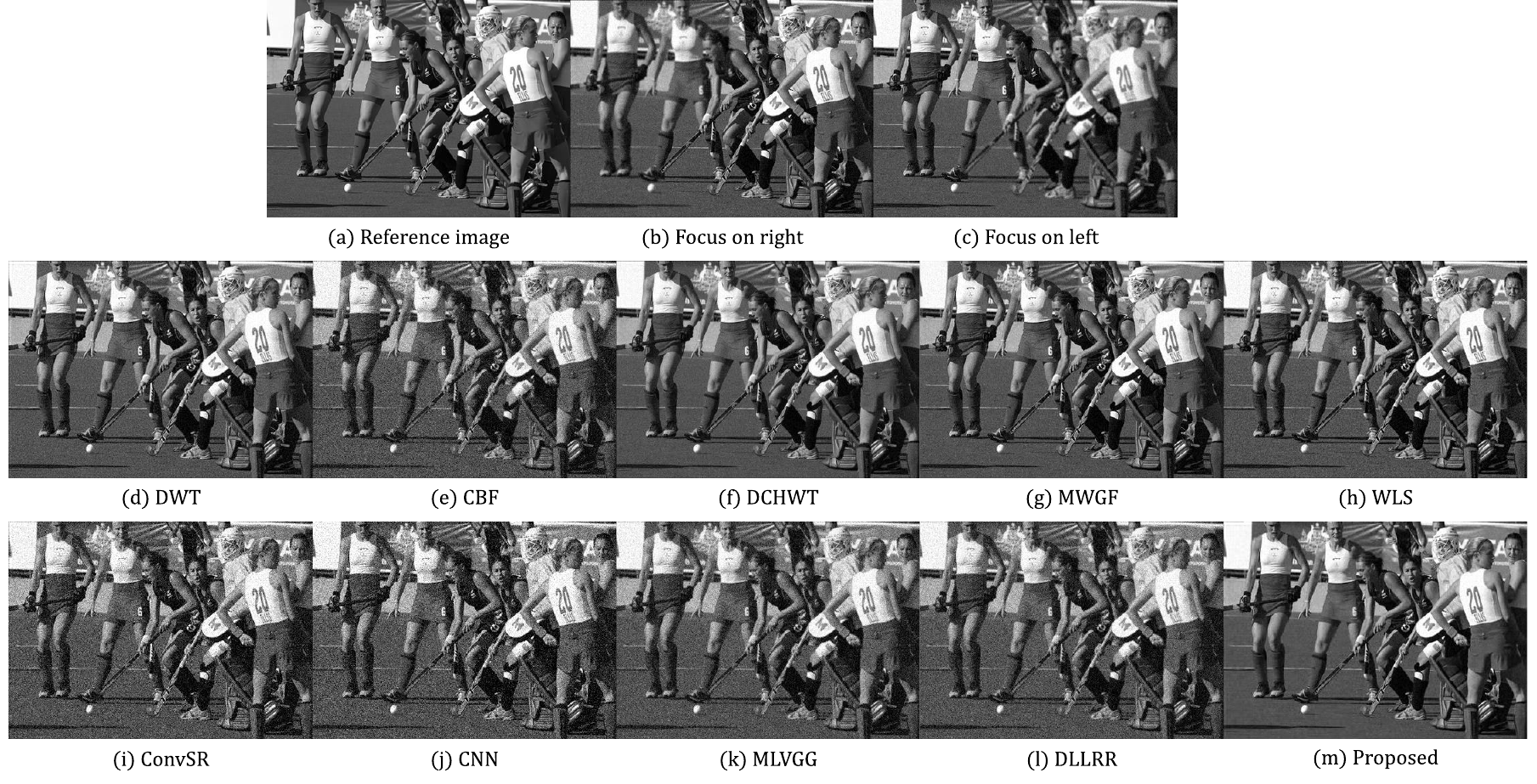}
\caption{The fused results obtained by fusion methods, and source images contain Gaussian noise($\sigma=0.001$). (a) Reference image(image 5); (b) Source image(focus on right); (c) Source image(focus on left); (d) DWT; (e) CBF; (f) DCHWT; (g) MWGF; (h) WLS; (i) ConvSR; (j) CNN; (k) MLVGG; (l) DLLRR; (m) Proposed}
\label{fig:fig7}
\end{figure*}

The fused results which are obtained by our method and other nine fusion methods are shown in Fig.\ref{fig:fig7}.

In Fig.\ref{fig:fig7}, source images contain Gaussian noise($\mu=0,\sigma=0.001$), this just shows an example for our fused results intuitively.

If fused image is more similar to reference image, then the fusion method has better fusion performance. As shown in Fig.\ref{fig:fig7}, the fused images obtained by CBF, ConvSR, CNN, MLVGG and DLLRR contain some noises, obviously. Compared with DWT, DCHWT, MWGF and WLS, the fused image obtained by our method is more similar to reference image, which means our fusion method can achieve better fusion performance.

The average values of RMSE, PSNR and SSIM for ten fused images obtained by existing methods and our fusion algorithm are shown in Table \ref{table:2}. And in our experiment, source images still contain three different types of noise. The best values are indicated in bold and the second-best values are denoted in red

\begin{table*}[ht]
\centering
\caption{\label{table:2}The average values of RMSE, PSNR and SSIM for ten fused images which obtained by fusion methods.}
\resizebox{\textwidth}{!}{
\begin{tabular}{*{13}{c}}
\midrule
 Noise &-- &Metrics &DWT &CBF &DCHWT &MWGF &WLS &ConvSR &CNN &MLVGG &DLLRR &Ours\\
\midrule
\multirow{12}*{{\shortstack{Gaussian\\ noise}}} &
\multirow{3}*{$\sigma=0.0005$} &
  RMSE	&0.03113 	&0.02537 	&0.02351 	&0.02307 	&0.03167 	&\textbf{0.02226} 	&\color{red}{0.02292} 	&0.02862 	&0.03497 	&0.02327\\ 
&&PSNR	&30.14548 	&31.93089 	&32.58992 	&\color{red}{32.74183} 	&30.00285 	&33.06375 	&\textbf{32.79989} 	&31.02855 	&29.14272 	&32.72188\\ 
&&SSIM	&0.81004 	&0.89162 	&0.88547 	&0.87179 	&0.81724 	&\color{red}{0.89293} 	&0.87405 	&0.88661 	&0.77964 	&\textbf{0.91120}\\ 
\cmidrule{2-13}
&\multirow{3}*{$\sigma=0.001$} &
  RMSE	&0.04131 	&0.03041 	&\color{red}{0.02981} 	&0.03188 	&0.04139 	&0.02999 	&0.03139 	&0.03268 	&0.03502 	&\textbf{0.02839}\\ 
&&PSNR	&27.68133 	&30.35320 	&\color{red}{30.52164} 	&29.92970 	&27.66704 	&30.46610 	&30.06572 	&29.81845 	&29.12999 	&\textbf{30.99987}\\ 
&&SSIM	&0.71738 	&0.83131 	&0.82113 	&0.79239 	&0.72592 	&0.81877 	&0.79725 	&\color{red}{0.83239} 	&0.77930 	&\textbf{0.87428}\\ 
\cmidrule{2-13}
&\multirow{3}*{$\sigma=0.005$} &
  RMSE	&0.08671 	&0.05597 	&0.05848 	&0.06917 	&0.08632 	&0.06474 	&0.06737 	&\color{red}{0.05463} 	&0.06891 	&\textbf{0.05059}\\ 
&&PSNR	&21.23926 	&25.04502 	&24.66211 	&23.20208 	&21.27921 	&23.77792 	&23.43194 	&\color{red}{25.27746} 	&23.23508 	&\textbf{26.02700}\\ 
&&SSIM	&0.45425 	&0.59962 	&0.58219 	&0.53231 	&0.46152 	&0.55600 	&0.54500 	&\color{red}{0.61366} 	&0.52955 	&\textbf{0.73387}\\ 
\cmidrule{2-13}
&\multirow{3}*{$\sigma=0.01$} &
  RMSE	&0.12028 	&0.07610 	&0.08002 	&0.09679 	&0.12007 	&0.09050 	&0.09342 	&\color{red}{0.07262} 	&0.09549 	&\textbf{0.05714}\\ 
&&PSNR	&18.39774 	&22.37396 	&21.93746 	&20.28483 	&18.41309 	&20.86844 	&20.59205 	&\color{red}{22.79417} 	&20.40204 	&\textbf{24.92995}\\ 
&&SSIM	&0.34680 	&0.48149 	&0.46482 	&0.41124 	&0.35154 	&0.43372 	&0.42838 	&\color{red}{0.49764} 	&0.41158 	&\textbf{0.64582}\\ 
\midrule

\multirow{6}*{{\shortstack{Salt \& pepper\\ noise}}} &
\multirow{3}*{$d=0.01$} &
  RMSE	&0.07673 	&0.07303 	&0.06396 	&0.06227 	&0.07765 	&0.06004 	&0.05557 	&\textbf{0.04743} 	&0.06898 	&\color{red}{0.04882}\\ 
&&PSNR	&22.30494 	&22.73372 	&23.88697 	&24.12445 	&22.20101 	&24.43670 	&25.10832 	&\textbf{26.52406} 	&23.22975 	&\color{red}{26.28470}\\ 
&&SSIM	&0.68092 	&0.69094 	&0.70344 	&0.76767 	&0.68251 	&0.76205 	&\textbf{0.80460} 	&\color{red}{0.77917} 	&0.71655 	&0.76127\\ 
\cmidrule{2-13}
&\multirow{3}*{$d=0.02$} &
  RMSE	&0.10665 	&0.09720 	&0.08391 	&0.08450 	&0.10836 	&0.08299 	&0.07782 	&\color{red}{0.06248} 	&0.09274 	&\textbf{0.05606}\\ 
&&PSNR	&19.44543 	&20.25157 	&21.52925 	&21.47467 	&19.30612 	&21.62228 	&22.18351 	&\color{red}{24.11548} 	&20.66027 	&\textbf{25.10640}\\ 
&&SSIM	&0.51530 	&0.53728 	&0.56410 	&0.63543 	&0.51534 	&0.62258 	&\color{red}{0.66816} 	&0.65694 	&0.57302 	&\textbf{0.69728}\\ 
\midrule

\multirow{3}*{Poisson noise} &
\multirow{3}*{--} &
  RMSE	&0.05077 	&\color{red}{0.03531} 	&0.03576 	&0.04018 	&0.05072 	&0.03745 	&0.03919 	&0.03694 	&0.04211 	&\textbf{0.03459}\\ 
&&PSNR	&25.90846 	&\color{red}{29.06021} 	&28.94820 	&27.94211 	&25.92070 	&28.55098 	&28.15376 	&28.72500 	&27.53885 	&\textbf{29.28811}\\ 
&&SSIM	&0.67047 	&0.78986 	&0.77762 	&0.74238 	&0.67943 	&0.77032 	&0.75138 	&\color{red}{0.79366} 	&0.73287 	&\textbf{0.83724}\\ 
\hline
\end{tabular}}
\end{table*}

From Table \ref{table:2}, when source images contain less noise, such as Gaussian noise with $\sigma=0.0005$ and Salt \& pepper noise with $d=0.01$, we don't have an obvious advantage. However, with the increase of noises, our method achieves almost best values in RMSE, PSNR and SSIM. This phenomenon indicates that our fusion algorithm can obtain better fusion performance when source images contain noise.

\section{Conclusions}\label{sec5}

In this paper, a novel noisy image fusion method based on low-rank representation has been proposed. In the DWT framework, the low frequency coefficients are fused by spatial frequency and choose-max scheme. And the high frequency coefficients are fused by low-rank representation and choose-max scheme. The experimental results show that the proposed method is more reasonable and more similar to original image. From the fused images and the values of RMSE, PSNR and SSIM, our method has better fusion performance compared with other methods when source images contain noise.

\section*{References}\label{References}

\begin{enumerate}
\item[{[1]}]	S. Li, X. Kang, L. Fang, J. Hu, and H. Yin, Pixel-level image fusion: A survey of the state of the art, Inf. Fusion, vol. 33, pp. 100-112, 2017.
\item[{[2]}]	A. Ben Hamza, Y. He, H. Krim, and A. Willsky, A Multiscale Approach to Pixel-level Image Fusion, Integr. Comput. Aided. Eng., vol. 12, pp. 135-146, 2005.
\item[{[3]}]	S. Yang, M. Wang, L. Jiao, R. Wu, and Z. Wang, Image fusion based on a new contourlet packet, Inf. Fusion, vol. 11, no. 2, pp. 78-84, 2010.
\item[{[4]}]	L. Wang, B. Li, and L. F. Tian, EGGDD: An explicit dependency model for multi-modal medical image fusion in shift-invariant shearlet transform domain, Inf. Fusion, vol. 19, no. 1, pp. 29-37, 2014.
\item[{[5]}]	Q. Zhang and B. long Guo, Multifocus image fusion using the nonsubsampled contourlet transform, Signal Processing, 2009.
\item[{[6]}]	Y. Zhang, X. Bai, and T. Wang, Boundary finding based multi-focus image fusion through multi-scale morphological focus-measure, Inf. Fusion, 2017.
\item[{[7]}]	Y. Liu, X. Chen, H. Peng, and Z. Wang, Multi-focus image fusion with a deep convolutional neural network, Inf. Fusion, vol. 36, pp. 191-207, 2017.
\item[{[8]}]	M. Nejati, S. Samavi, and S. Shirani, Multi-focus image fusion using dictionary-based sparse representation, Inf. Fusion, vol. 25, pp. 72-84, 2015.
\item[{[9]}]	H. Yin, Y. Li, Y. Chai, Z. Liu, and Z. Zhu, A novel sparse-representation-based multi-focus image fusion approach, Neurocomputing, vol. 216, pp. 216-229, 2016.
\item[{[10]}]	Y. Liu, X. Chen, R. K. Ward, and J. Wang, Image Fusion with Convolutional Sparse Representation, IEEE Signal Process. Lett., vol. 23, no. 12, pp. 1882-1886, 2016.
\item[{[11]}]	H. Li and X.-J. Wu, Multi-focus Image Fusion using dictionary learning and Low-Rank Representation, in Image and Graphics. ICIG 2017. Lecture Notes in Computer Science, vol 10666. Springer, Cham., 2017, pp. 675-686.
\item[{[12]}]	G. Liu, Z. Lin, and Y. Yu, Robust Subspace Segmentation by Low-Rank Representation, in Proceedings of the 27th International Conference on Machine Learning, 2010, pp. 663-670.
\item[{[13]}]	B. K. Shreyamsha Kumar, Image fusion based on pixel significance using cross bilateral filter, Signal, Image Video Process., 2015.
\item[{[14]}]	B. K. Shreyamsha Kumar, Multifocus and multispectral image fusion based on pixel significance using discrete cosine harmonic wavelet transform, Signal, Image Video Process., vol. 7, no. 6, pp. 1125-1143, 2013.
\item[{[15]}]	Z. Zhou, S. Li, and B. Wang, Multi-scale weighted gradient-based fusion for multi-focus images, Inf. Fusion, vol. 20, no. 1, pp. 60-72, 2014.
\item[{[16]}]	J. Ma, Z. Zhou, B. Wang, and H. Zong, Infrared and visible image fusion based on visual saliency map and weighted least square optimization, Infrared Phys. Technol., vol. 82, pp. 8-17, 2017.
\item[{[17]}]	H. Li, X.-J. Wu, and J. Kittler, Infrared and Visible Image Fusion using a Deep Learning Framework, in arXiv preprint arXiv:1804.06992, 2018.

\item[{[18]}] Luo X, Zhang Z, Wu X. A novel algorithm of remote sensing image fusion based on shift-invariant Shearlet transform and regional selection[J]. AEU-International Journal of Electronics and Communications, 2016, 70(2): 186-197.
\item[{[19]}] Zheng Y J, Yang J Y, Yang J, et al. Nearest neighbour line nonparametric discriminant analysis for feature extraction[J]. Electronics Letters, 2006, 42(12): 679-680.
\item[{[20]}] Sun J, Fang W, Wu X J, et al. Quantum-behaved particle swarm optimization: principles and applications[J]. 2011.
\item[{[21]}] Wang M, Wang S T, Wu X J. Initial results on fuzzy morphological associative memories[J]. Acta Electronica Sinica, 2003, 31(5): 690-693.
\item[{[22]}] Li H, Wu X J. DenseFuse: A fusion approach to infrared and visible images[J]. IEEE Transactions on Image Processing, 2018, 28(5): 2614-2623.

\end{enumerate}

\end{document}